\documentclass[runningheads]{llncs}
\usepackage{graphicx}
\usepackage{multirow}

\usepackage{stmaryrd}
\usepackage{amsmath}
\usepackage{amsfonts}

\SetSymbolFont{stmry}{bold}{U}{stmry}{m}{n}

\usepackage{ifthen}
\usepackage{algorithm}
\usepackage{algorithmic}

\newcommand{\mv}{\boldsymbol{m}}
\newcommand{\cov}{\boldsymbol{C}}
\newcommand{\context}{\boldsymbol{\alpha}}

\newcommand{\x}{\boldsymbol{x}}
\newcommand{\y}{\boldsymbol{y}}
\newcommand{\z}{\boldsymbol{z}}

\newcommand{\E}{\mathbb{E}}

\newcommand{\R}{\mathbb{R}}

\newcommand{\Tr}{\mathrm{Tr}}
\newcommand{\N}{\mathcal{N}}

\newcommand{\argmin}{\mathop{\rm arg~min}\limits}

\newcommand{\T}{\mathrm{T}}

\newcommand{\I}{\mathbf{I}}

\newcommand{\agl}[1]{\langle {#1} \rangle}
\newcommand{\p}{\boldsymbol{p}}

\usepackage{xcolor}
\usepackage{ifthen}
\newcommand{\markupdraft}[2]{
    \ifthenelse{\equal{#1}{display}}{#2}{}
    \ifthenelse{\equal{#1}{color}}{\color{#2}}{}
}
\newcommand{\notecolored}[3][]{\markupdraft{display}{{\color{#2}\noindent[Note (#1): #3]}}}
\newcommand{\newcolored}[3][]{{\markupdraft{color}{#2}#3}
    \ifthenelse{\equal{#1}{}}{}{\markupdraft{display}{{\color{yellow!70!black}[#1]}}}}
\newcommand{\del}[2][]{{\markupdraft{display}{{\color{orange}[removed: ``#2''[#1]]}}}} 
\newcommand{\new}[2][]{\newcolored[#1]{blue}{#2}}
\renewcommand{\note}[2][]{\notecolored[#1]{green}{#2}}

\renewcommand{\del}[2]{}  
\renewcommand{\markupdraft}[2]{}  

\usepackage[subrefformat=parens]{subcaption}
\begin{document}
\title{Warm Starting of CMA-ES for Contextual Optimization Problems}
%
%

\author{Yuta Sekino
\and Kento Uchida 
\and Shinichi Shirakawa
}

\authorrunning{Y. Sekino et al.}

\institute{Yokohama National University, Yokohama, Japan\\
\email{
sekino-yuta-cs@ynu.jp,
uchida-kento-fz@ynu.ac.jp, 
shirakawa-shinichi-bg@ynu.ac.jp}
}


%
\maketitle              
\begin{abstract}
\note{The abstract should briefly summarize the contents of the paper in 15--250 words.}
Several practical applications of evolutionary computation possess objective functions that receive the design variables and externally given parameters. Such problems are termed contextual optimization problems. These problems require finding the optimal solutions corresponding to the given context vectors. Existing contextual optimization methods train a policy model to predict the optimal solution from context vectors. However, the performance of such models is limited by their representation ability. By contrast, warm starting methods have been used to initialize evolutionary algorithms on a given problem using the optimization results on similar problems. Because warm starting methods do not consider the context vectors, their performances can be improved on contextual optimization problems. Herein, we propose a covariance matrix adaptation evolution strategy with contextual warm starting (CMA-ES-CWS) to efficiently optimize the contextual optimization problem with a given context vector. The CMA-ES-CWS utilizes the optimization results of past context vectors to train the multivariate Gaussian process regression. Subsequently, the CMA-ES-CWS performs warm starting for a given context vector by initializing the search distribution using posterior distribution of the Gaussian process regression. The results of the numerical simulation suggest that CMA-ES-CWS outperforms the existing contextual optimization and warm starting methods.

\end{abstract}
%
%
%

\section{Introduction}
In practical applications of evolutionary algorithms, the evaluation value of the given objective function is often determined using the design variables and externally given parameters.
These optimization problems are referred to as contextual optimization problems and require finding the optimal solution
corresponding to the given external parameter, called context vector.
Contextual optimization problems arise in many applications, such as the optimization of the controller model of a robot for target motion~\cite{application3:robotics,application2:robotics} and nuclear fusion control using plasma states as context vectors~\cite{application1:nuclear}.
For example, in the controller model of robot locomotion task~\cite{application3:robotics}, the target locomotion speed and the locomotion direction are possible choices of the context vector, and the objective of optimization is to design a controller model that realizes the locomotion of the robot with the target speed.

Various contextual optimization methods have been proposed to efficiently optimize these contextual optimization problems~\cite{abdolmaleki:2016,ccmaes,creps}. 
These methods solve contextual optimization problems by training the policy model that receives the context vector and predicts the corresponding optimal solution.
To reduce the training cost of the policy model in contextual optimizations, several studies~\cite{ccmaes,Abdolmaleki:2019} have focused on the update rules of the covariance matrix adaptation evolution strategy (CMA-ES)~\cite{hansen:1996:ec}, which employs a multivariate Gaussian distribution as the sampling distribution and iteratively updates the distribution parameters.
The contextual CMA-ES~\cite{ccmaes} extends the update rules of the CMA-ES to update the policy model.
Notably, because many contextual optimization methods use linear models as the policy model, their ability to predict the optimal solution is limited.

Meanwhile, warm starting methods have been used to start the optimization of evolutionary algorithms from a good state on a given problem.
The warm starting methods initialize the evolutionary algorithm using the optimization results on similar problems~\cite{wscmaes,wstpe}.
For example, the warm starting CMA-ES (WS-CMA-ES) initializes the multivariate Gaussian distribution using the set of superior solutions achieved in the optimization of a given similar task. 
However, the existing warm starting methods do not consider the context vectors.
We are of the view that the performance of warm starting methods on contextual optimization problems can be improved using the information of context vectors. 
A related idea was introduced in the contextual bandit problems~\cite{arrowcb}, which utilize a dataset consisting of a set of context vectors provided in advance by experts and a set of reward vectors for each action.

In this study, we propose the CMA-ES with contextual warm starting (CMA-ES-CWS) to efficiently obtain the optimal solution corresponding to a given context vector.
CMA-ES-CWS requires the best solutions corresponding to some context vectors achieved in past optimizations.
Given the optimization results for past context vectors, CMA-ES-CWS trains the multi-output Gaussian process regression (GPR) to predict the optimal solution from the context vector. 
Next, CMA-ES-CWS performs warm starting for a newly obtained context vector by initializing the search distribution using the predictive distribution of the GPR.

We evaluate the performance of the CMA-ES-CWS using the benchmark functions and control task of the robot arm.
In the experiment with benchmark functions, we transform the search space of existing benchmark functions based on the context vector.
Experimental results show that, with nonlinear and noisy transformations, CMA-ES-CWS can obtain the optimal solution corresponding to the given context vector more efficiently than the contextual CMA-ES and WS-CMA-ES.
In addition, we confirm that the performance of CMA-ES-CWS is improved by increasing the amount of past optimization results.
Further, we evaluate the performance of CMA-ES-CWS in robot control tasks.

\section{Preliminaries}

\subsection{CMA-ES}

The CMA-ES is a probabilistic model-based black-box continuous optimization method using the multivariate Gaussian distribution $\N \left( \mv, \sigma^2 \cov \right)$
which is parameterized by mean vector $\mv^{(t)} \in \R^N$, covariance matrix $\cov^{(t)} \in \R^{N \times N}$, and step-size $\sigma^{(t)} \in \R_{>0}$.

In each iteration, the CMA-ES generates $\lambda$ solutions $\x^{\agl{1}}, \cdots, \x^{\agl{\lambda}}$ from the Gaussian distribution as
\begin{align}
    \z^{\agl{i}} \sim \N(\mathbf{0}, \I) \quad \text{,} \quad
    \y^{\agl{i}} = \sqrt{\cov^{(t)}} \z^{\agl{i}} \quad \text{and} \quad
    \x^{\agl{i}} = \mv^{(t)} + \sigma^{(t)} \y^{\agl{i}} \enspace.
\end{align}
Next, the CMA-ES computes the ranking of the solutions on the objective function.
We denote the index of the $i$-th best solution as $i\!:\!\lambda$.

Next, the CMA-ES updates two evolution paths: $\p_c \in \R^{N}$ and $\p_\sigma \in \R^{N}$.
These evolution paths are initialized as $\p_\sigma^{(0)} = \p_c^{(0)} = \mathbf{0}$.
The CMA-ES computes two weighted sums, namely, $\Delta_{\z} = \sum_{i = 1}^{\mu} w_i \z^{\agl{i:\lambda}}$ and $\Delta_{\y} = \sum_{i = 1}^{\mu} w_i \y^{\agl{i:\lambda}}$ of $\mu$ best solutions, to update the evolution paths as
\begin{align}
    \p_\sigma^{(t+1)} &= (1 - c_\sigma) \p_\sigma^{(t)} + \sqrt{ c_\sigma (2 - c_\sigma) \mu_\mathrm{eff} } \cdot \Delta_{\z} \\
    \p_c^{(t+1)} &= (1 - c_c) \p_c^{(t)} + h_\sigma^{(t+1)} \sqrt{ c_c (2 - c_c) \mu_\mathrm{eff} } \cdot \Delta_{\y} \enspace,
\end{align}
where $c_\sigma, c_c \in (0,1]$ are accumulation factors, and $\mu_\mathrm{eff} = (\sum_{i=1}^\mu w_i^2)^{-1}$ is the variance effective selection mass.
The Heaviside function $h_\sigma \in \{0, 1\}$ is set to $h_\sigma = 1$ when it satisfies
\begin{align}
    \frac{\| \p_\sigma^{(t+1)} \| }{\sqrt{1 - (1 - c_\sigma)^{2 (t + 1)}}} < \left( 1.4 + \frac{2}{N+1} \right) \chi_N \enspace,
\end{align}
where $\chi_{N} = \sqrt{N} \left( 1 - \frac{1}{4 N} + \frac{1}{21 N^2} \right)$ denotes the approximated value of $\E[ \| \N(\mathbf{0}, \I) \|]$.

Finally, the CMA-ES updates the distribution parameters.
The update rule of the mean vector is given as
\begin{align}
    \mv^{(t+1)} &= \mv^{(t)} + c_m \sigma^{(t)} \Delta_{\y} \enspace,
    \label{eq:cma:update-mv}
\end{align}
where the learning rate $c_m \in (0,1]$ is usually set as $c_m = 1$.
The update of the covariance matrix consists of the rank-$\mu$ update and the rank-one update.
The rank-$\mu$ update uses the weighted sum of the $\mu$ best solutions, whereas the rank-one update uses the evolution path $\p_c$.
The update rule of the covariance matrix is given by
\begin{multline}
        \cov^{(t+1)} = \left(1 + (1 - h_\sigma^{(t+1)}) c_1 c_c (2 - c_c) \right) \cov^{(t)} \\
        + \underbrace{ c_\mu \sum_{i = 1}^{\mu} w_i \left( \y^{\agl{i:\lambda}}  \left( \y^{\agl{i:\lambda}} \right)^\T - \cov^{(t)} \right) }_{\text{rank-$\mu$ update}} 
        + \underbrace{ c_1 \left( \p_c^{(t+1)} \left( \p_c^{(t+1)} \right)^\T - \cov^{(t)} \right) }_{\text{rank-one update}} 
        \enspace,
    \label{eq:cma:update-cov} 
\end{multline}
where $c_\mu, c_1 \in (0,1]$ denote the learning rates.
The update rule of the step-size is given as
\begin{align}
    \sigma^{(t+1)} &= \sigma^{(t)} \exp \left( \frac{c_\sigma}{d_\sigma} \left( \frac{ \| \p_\sigma^{(t+1)} \| }{\chi_N} - 1 \right) \right)
    \label{eq:cma:update-stepsize}   
    \enspace,
\end{align}
where $d_\sigma \in \R_{>0}$ denotes the damping factor.
The CMA-ES has well-tuned recommended settings for each hyperparameter, as shown in~\cite{hansen:2017:arxiv,hansen:2014:book}.

The initial distribution parameters $\mv^{(0)}, \cov^{(0)},$ and $\sigma^{(0)}$ significantly influence the optimization performance.
However, because the precise general-purpose method to determine these important hyperparameters does not exist, they are manually set (or sometimes manually tuned) for the target problem.

\subsection{Multi-output Gaussian Process Regression}

The GPR is a non-parametric regression method, which expresses the prediction using the posterior distribution of the function following a Gaussian process.
The Gaussian process is a function $g$ of the distribution whose evaluation values at arbitrary $n$ points $\mathcal{D}_n = \{ \x_i \}_{i=1}^n$ follow the multivariate Gaussian distribution $\N(\mathbf{0}, \boldsymbol{K})$.

The prediction of GPR uses the conditional distribution of the multivariate Gaussian distribution.
We denote sample $\x$, mean vector $\boldsymbol{\mu}$, and covariance matrix ${\bf\Sigma}$ as
\begin{align}
    \x = \left( \begin{array}{c}
         \x_a \\ \x_b
    \end{array} \right), \quad
    \boldsymbol{\mu} = \left( \begin{array}{c}
         \boldsymbol{\mu}_a \\ \boldsymbol{\mu}_b
    \end{array} \right), \quad
    {\bf\Sigma} = \left( \begin{array}{cc}
        {\bf\Sigma}_{a,a} & {\bf\Sigma}_{a,b} \\
        {\bf\Sigma}_{b,a} & {\bf\Sigma}_{b,b}
    \end{array} \right) \enspace.
\end{align}
Then, the distribution of $\x_b$ conditioned on $\x_a$ is given by a multivariate Gaussian distribution with mean vector $\boldsymbol{\mu}_{b|a}$ and covariance matrix ${\bf\Sigma}_{b|a}$ as
\begin{align}
    \boldsymbol{\mu}_{b|a} = \boldsymbol{\mu}_{b} + {\bf\Sigma}_{b,a} {\bf\Sigma}_{b,b}^{-1} (\x_b - \boldsymbol{\mu}_b) 
    \quad\text{and}\quad
    {\bf\Sigma}_{b|a} = {\bf\Sigma}_{b,b} - {\bf\Sigma}_{b,a} {\bf\Sigma}_{a,a}^{-1} {\bf\Sigma}_{a,b} \enspace.
    \label{eq:gpr:conditional} 
\end{align}

Here, we consider the multi-output GPR that predicts a function $g$ with $L$ outputs.
In this case, the vector $\boldsymbol{g}_n = (g(\x_1)^\T, \cdots, g(\x_n)^\T)^\T \in \R^{L n}$
that consists of evaluation values at points in $\mathcal{D}_n$ follows the $(L n)$-dimensional multivariate Gaussian distribution whose covariance matrix is given by the Gram matrix $\boldsymbol{K}_L(\mathcal{D}_n) \in \R^{(L n) \times (L n)}$.
There are several methods to construct the Gram matrix for multi-output GPR. We use the linear model of coregionalization (LMC)~\cite{lmc}.
The LMC introduces $Q$ matrices $\boldsymbol{B}_1, \cdots, \boldsymbol{B}_Q \in \R^{L \times L}$ and $Q$ kernels $k_1, \cdots, k_Q$ and computes the Gram matrix as
\begin{align}
    \boldsymbol{K}_L(\mathcal{D}_n) = \sum^Q_{q=1} \boldsymbol{K}(\mathcal{D}_n; k_q) \otimes \boldsymbol{B}_q = \left( \begin{array}{cc}
        \boldsymbol{K}_L(\mathcal{D}_{n-1}) & \boldsymbol{K}_\ast \\
        \boldsymbol{K}_\ast^\T & \boldsymbol{K}_{\ast\ast}
    \end{array} \right) \enspace,
\end{align}
where the operation $\otimes$ is the Kronecker product, and $\boldsymbol{K}_{\ast} \in \R^{Ln \times L}$ and $\boldsymbol{K}_{\ast\ast} \in \R^{L \times L}$ are components of the decomposition of the Gram matrix.
The covariance matrix $\boldsymbol{K}(\mathcal{D}_n; k_q)$ using the kernel function $k_q: \R^N \times \R^N \to \R$ is given by
\begin{align}
     \boldsymbol{K}(\mathcal{D}_n; k_q) &= \left( \begin{array}{ccc}
        k_q(\boldsymbol{x}_1, \boldsymbol{x}_1) & \cdots & k_q(\boldsymbol{x}_n, \boldsymbol{x}_1) \\
        \vdots && \vdots \\
        k_q(\boldsymbol{x}_1, \boldsymbol{x}_n) & \cdots & k_q(\boldsymbol{x}_n, \boldsymbol{x}_n) 
    \end{array} \right) \enspace.
    \label{eq:gram:multi}
\end{align}
The LMC computes the predictive distribution of the function output at $\x_n$ as a $L$-dimensional multivariate Gaussian distribution $\N(\boldsymbol{\mu}(\boldsymbol{x}_n ), {\bf\Sigma}(\boldsymbol{x}_n ))$, where
\begin{align}
    \boldsymbol{\mu}(\boldsymbol{x}_n ) &= \boldsymbol{K}_{\ast}^\T \left( \boldsymbol{K}_L(\mathcal{D}_{n-1}) \right)^{-1} \boldsymbol{g}_{n-1}  
    \label{eq:gpr:conditional:multmean}
    \\
    {\bf\Sigma}(\boldsymbol{x}_n ) &= \boldsymbol{K}_{\ast \ast} - \boldsymbol{K}_{\ast}^\T \left( \boldsymbol{K}_L(\mathcal{D}_{n-1}) \right)^{-1} \boldsymbol{K}_{\ast} 
    \label{eq:gpr:conditional:multsigma}
    \enspace.
\end{align}

In the prediction of LMC, the $d$-th element of the function output $g(\x_n)$ can be interpreted as the weighted sum of $QR$ Gaussian processes with $Q$ kernels as
\begin{align}
    (g(\x))_d = \sum^{Q}_{q=1} \sum^{R}_{r=1} a^r_{d,q} u^r_q(\x) \enspace,
\end{align}
where $u^r_q : \R^N \to \R$ is the sample from the Gaussian process with kernel $k_q$.
Coefficient $a^r_{d,q} \in \R$ is set such that as to satisfy $(\boldsymbol{B}_q)_{d,d'} = \sum^{R}_{r=1} a^r_{d,q} a^r_{d',q}$.

\section{Problem Definition} \label{sec:problem}
We consider a contextual optimization problem whose objective function \new{$f(\x, \context)$} is determined by context vector $\context \in \R^{N_\alpha}$.
The objective of the contextual optimization problem is to obtain an optimal solution corresponding to the given context vector.

\paragraph{Objective of existing contextual optimization methods}:
The existing contextual optimization methods~\cite{cps,acps,cps2} aim to train the policy model $\pi_{\boldsymbol{w}}$ to predict the optimal solution $\x^\ast(\context)$ from the context vector $\context$.
Their target is to obtain the optimal parameter $\boldsymbol{w}^\ast$ of policy model that minimizes the expected objective function value under the distribution $p_\alpha$ of the context vector as
\begin{align}
    \boldsymbol{w}^\ast = \argmin_{\boldsymbol{w} \in \mathcal{W}} \int_{\x} \int_{\context} p_\alpha(\context) \pi_{\boldsymbol{w}}( \x \mid \context) f(\x; \context) \mathrm{d}\context \mathrm{d}\x \enspace.
\end{align}
There are different scenarios of the training in contextual optimization: one scenario is contextual policy search~\cite{cps,cps2} where the optimizer receives the context vectors stochastically generated, and another is the active contextual policy search~\cite{acps}, where the optimizer can determine the context vector.

\paragraph{Objective of this study}:
Different from the objective of the existing contextual optimization methods, the objective of this study is to achieve efficient optimization of objective function $f(\x ; \context_\mathrm{new})$ for $\x$ after we receive a target context vector $\context_\mathrm{new}$.
The optimal solution corresponding to a target context vector $\context_\mathrm{new}$ is formulated as
\begin{align}
    \x^\ast(\context_\mathrm{new}) &= \argmin_{\x \in \R^N} f(\x ; \context_\mathrm{new}) \enspace. \label{eq:problem}
\end{align}
In this study, the domain of design variables $\x$ is the continuous space $\R^N$. 
We assume that $M_\mathrm{prev}$ best solutions $\x^\mathrm{best}_1, \cdots, \x^\mathrm{best}_{M_\mathrm{prev}}$ for context vectors $\context_1, \cdots, \context_{M_\mathrm{prev}}$ are obtained in advance.

\begin{algorithm*}[t]
\caption{CMA-ES with contextual warm-starting (CMA-ES-CWS)}
\label{alg:proposed}
\begin{algorithmic}[1]
\REQUIRE Pairs of context and best solution found previously $\mathcal{D} = \{ (\context_1, \x^\mathrm{best}_1), \cdots, (\context_{M_\mathrm{prev}}, \x^\mathrm{best}_{M_\mathrm{prev}}) \}$ 
\REQUIRE Target context vector $\context_\mathrm{new}$
\STATE Compute the mean vector $\boldsymbol{\mu}(\context_\mathrm{new})$ and covariance matrix ${\bf\Sigma}(\context_\mathrm{new})$ of the predictive distribution using \eqref{eq:gpr:conditional:multmean} and \eqref{eq:gpr:conditional:multsigma}.
\STATE Set initial distribution parameters to $\mv^{(0)} = \boldsymbol{\mu}(\context_\mathrm{new})$, $\cov^{(0)} = \I$, and $\sigma^{(0)} = \mathrm{clip}( \sqrt{\Tr ({\bf\Sigma}(\context_\mathrm{new})) / N}, \sigma_{\min},\sigma_{\max})$.
\STATE Run CMA-ES with the initial distribution parameters $\mv^{(0)}, \sigma^{(0)}, \cov^{(0)}$.
\end{algorithmic}
\end{algorithm*}
%
%
%

\section{Proposed Method: CMA-ES-CWS}
In this study, we propose CMA-ES with contextual warm-starting (CMA-ES-CWS), which introduces warm starting for contextual optimization problems to CMA-ES.
The CMA-ES-CWS utilizes optimization results for past context vectors to train the multi-output GPR. Subsequently, it performs warm starting for a given context vector by initializing the sampling distribution using the predictive distribution of the GPR.
Algorithm~\ref{alg:proposed} shows the pseudocode of CMA-ES-CWS.

\subsection{Predictive Distribution for Optimal Solution}
The CMA-ES-CWS utilizes $M_\mathrm{prev}$ pairs of context vector and optimization result $\mathcal{D} = \{ (\context_1, \x^\mathrm{best}_1), \cdots (\context_{M_\mathrm{prev}}, \x^\mathrm{best}_{M_\mathrm{prev}}) \}$ to predict the distribution of the optimal solution $\x^\ast_\mathrm{new}$ corresponding to a newly given context vector $\context_\mathrm{new}$.
The multi-output GPR represents the predictive distribution using a multivariate Gaussian distribution as
\begin{align}
    p( \x^\ast_\mathrm{new} \mid \context_\mathrm{new}, \mathcal{D} ) = \N( \boldsymbol{\mu}(\context_\mathrm{new}), {\bf\Sigma}(\context_\mathrm{new})) \enspace.
\end{align}
We use LMC to compute the predictive distribution.
We apply three kernels for the computation of the Gram matrix in \eqref{eq:gram:multi}: linear kernel $k_1$, radial basis function (RBF)  kernel $k_2$, and Matern 5/2 kernel $k_3$ as 
\begin{align}
    k_1(\x, \x') &= \sigma_1^2 \x^\T \x' \\
    k_2(\x, \x') &= \sigma_2^2 \exp\left( - \frac{1}{2} (r_2(\x, \x'))^2 \right) \\
    k_3(\x, \x') &= \sigma_3^2 \left( 1 + \sqrt{5} r_3(\x, \x') + \frac{5}{3} (r_3(\x, \x'))^2 \right)  \exp\left( - \sqrt{5} r_3(\x, \x') \right) \enspace,
\end{align}
where $\sigma_q \in \R_{>0}$ and $\ell_{q,i} \in \R_{>0}$ are hyperparameters.
The RBF and Matern 5/2 kernels are functions of $r_q \in \R_{>0}$ for $q=2,3$ defined as $r_q(\x, \x') = \sqrt{ \sum^N_{i=1}{( x_i - x'_i )^2} / { \ell_{q,i}^2 } }$.
Matrix $\boldsymbol{B}_q$ is determined using hyperparameters $\boldsymbol{a}^r_q \in \R^N$ and $\kappa \in \R_{>0}$ as
\begin{align}
    \boldsymbol{B}_q = \sum_{r=1}^R \boldsymbol{a}^r_q (\boldsymbol{a}^r_q)^\T + \kappa \I \enspace.
\end{align}
We set $R=1$.
These hyperparameters of each kernel are optimized through marginal likelihood maximization.
\footnote{\texttt{Gpy 1.10.0}~\cite{gpy} was used to implement the proposed method.}

\subsection{Warm Starting Using Predictive Distribution}
The CMA-ES-CWS uses the mean vector $\boldsymbol{\mu}(\context_\mathrm{new})$ and covariance matrix ${\bf\Sigma}(\context_\mathrm{new})$ of the predictive distribution to obtain the initial values of the probability distribution parameters as follows:
\begin{align}
    \mv^{(0)} &= \boldsymbol{\mu}(\context_\mathrm{new}) \\
    \sigma^{(0)} &= \mathrm{clip}\left( \sqrt{ \frac{\Tr ({\bf\Sigma}(\context_\mathrm{new}))}{N} },\sigma_{\min},\sigma_{\max} \right) \label{eq:proposed:sigma} \\
    \cov^{(0)} &= \I
\end{align}
where $\mathrm{clip}(a,b,c) = \min\{ \max\{a, b\}, c \}$ and $\sigma_{\min}=10^{-2}, \sigma_{\max}=2$.
This clipping prevents the inefficient optimization caused by a too small initial step-size.
The CMA-ES-CWS uses these initial values to search for the optimal solution $\x^\ast(\context_\mathrm{new})$ through CMA-ES.

Note that CMA-ES-CWS does not use the covariance matrix ${\bf\Sigma}(\context_\mathrm{new})$ of the predictive distribution to initialize the covariance matrix $\cov$ in CMA-ES.
This is because the appropriate distribution shape in CMA-ES depends on the functional shape of the objective function rather than the location of the optimal solution, whereas the predictive distribution given by the multiple-output GPR uses only information of the best solution, not the functional shape.

\section{Experiment Using Benchmark Functions}
\label{sec:experiment:benchmark}
In this section, we present the result of the performance evaluation of CMA-ES-CWS using benchmark functions.
In Section~\ref{sec:experiment:benchmark:comparison}, we compare the CMA-ES-CWS with existing methods using benchmark functions.
In Section~\ref{sec:experiment:benchmark:ablation}, we verify the relationship between the number of past optimization results $M_\mathrm{prev}$ and performance of CMA-ES-CWS.

\subsection{Comparative Methods}
\paragraph{Contextual CMA-ES:}
Contextual CMA-ES~\cite{ccmaes} is a contextual optimization method that uses the update rules of CMA-ES.
As a traditional problem setting of contextual optimization described in Section~\ref{sec:problem}, contextual CMA-ES assumes that the context vector is given randomly before generating a solution.

The contextual CMA-ES trains a policy model to predict the optimal solution corresponding to the context vector $\context$.
It predicts the optimal solution using the linear model $\mv^{(t)}( \context ) = \boldsymbol{A}^{(t)} \varphi(\context)$ with parameter $\boldsymbol{A} \in \R^{N \times N_{\varphi}}$,
where $\varphi: N_\alpha \to N_\varphi$ determines the context features.
In our experiment, we set $\varphi(\context) = (\context^\T, 1)^\T$ and $N_\varphi = N+1$, in accordance with the settings in reference~\cite{ccmaes}.
The contextual CMA-ES acquires the policy model by repeatedly generating samples from the Gaussian distribution $\N( \mv( \context ), \sigma^2 \cov )$ and updating the parameters $\boldsymbol{A}, \sigma,$ and $\cov$.
The mean vector of Gaussian distribution is gained from the predictions of the linear model.

\paragraph{WS-CMA-ES:}
Warm starting CMA-ES (WS-CMA-ES)~\cite{wscmaes} uses the evaluated solution set obtained in the optimization of a similar task to initialize CMA-ES for a newly given target task.
It uses the best $K_\gamma = \lfloor \gamma K \rfloor$ solutions $ \x_{1}, \cdots, \x_{K_\gamma} $ out of the $K$ solutions obtained in a similar task to compute the promising distribution for the target task as
\begin{align}
    p(\x) = \frac{1}{K_\gamma} \sum^{K_\gamma}_{i=1} \N( \x_{i}, \alpha^2 \I ) \enspace,
\end{align}
where $\gamma=0.1$ and $\alpha=0.1$.
The WS-CMA-ES optimizes the target task from a Gaussian distribution that has the smallest Kullback–Leibler divergence from the promising distribution.
The mean vector and covariance matrix of this Gaussian distribution are analytically given as follows:
\begin{align}
    \mv^\ast = \frac{1}{K_\gamma} \sum^{K_\gamma}_{i=1} \x_i \quad \text{and} \quad
    {\bf\Sigma}^\ast = \alpha^2 \I + \frac{1}{K_\gamma} \sum^{K_\gamma}_{i=1} (\x_i - \mv^\ast) (\x_i - \mv^\ast)^\T 
\end{align}
The WS-CMA-ES initializes the Gaussian distribution to satisfy $\mv^{(0)} = \mv^\ast$ and $(\sigma^{(0)})^2 \cov^{(0)} = {\bf\Sigma}^\ast$.
Note that WS-CMA-ES does not consider the existence of context vectors.

\begin{table*}[!t]
\centering
\caption{Definitions of benchmark functions}
\label{table:Benchmark}
\begin{tabular}{cll} \hline
   No. & Name & Definition \\ \hline\hline
   1 & Sphere & $f_1(\boldsymbol{y})=\sum_{i=1}^{N} y_i^2$ \\
   2 & Rosenbrock & 
   $f_2(\boldsymbol{y}) = \sum_{i=1}^{N-1} \left(100(y_{i+1}-y_{i}^2)^2+(1-y_{i})^2\right)$ \\
   3 & Easom & 
   $f_3(\boldsymbol{y})=-\cos(y_1)\cos(y_2)\exp(-((y_1-\pi)^2+(y_2-\pi)^2)) + 1$ \\ \hline
\end{tabular}
\end{table*}
%
%
%

\subsection{Experimental Setting}\label{sec:experiment:benchmark:setting}
The experimental setting partially followed the setting used in~\cite{ccmaes}.
We used the benchmark functions $f(\x; \context) = f_i(\phi_j(\x; \context))$ constructed by applying a context vector dependent transformation $\phi_j(\x, \alpha)$ to the existing black-box continuous benchmark function $f_i$.
Table~\ref{table:Benchmark} shows the definitions of benchmark functions.

The sphere function is a unimodal benchmark function that is easy to optimize.
The Rosenbrock function is an ill-scale and non-separable benchmark function that requires covariance matrix adaptation to fit the function landscape.
The Easom function has many shallow local optima and requires starting optimization with initial distribution around the optimal solution.
In this experiment, we set the number of dimensions to $N=20$ for the sphere and Rosenbrock functions, and $N=2$ for the Easom function.
We used three transformations depending on the context vector as follows:


\begin{itemize}
    \item Linear Shift: $\phi_1(\boldsymbol{x}; \context)=\boldsymbol{x}-\boldsymbol{G}\boldsymbol{\alpha}$ 
    \item Nonlinear Shift: $\phi_2(\boldsymbol{x}; \context)=\boldsymbol{x}-\boldsymbol{G} (\boldsymbol{\alpha} \circ \boldsymbol{\alpha})$ 
    \item Noisy Shift: $\phi_3(\boldsymbol{x}; \context)=\boldsymbol{x}-\boldsymbol{G}\boldsymbol{\alpha} + \epsilon^2 \mathcal{N}$
\end{itemize}

Each element of constant matrix $\boldsymbol{G} \in \R^{N \times N_\alpha}$ was assigned according to the standard normal distribution $\mathcal{N}(0,1)$ and shared in the optimizations for the target and previously appeared $M_\mathrm{prev}$ context vectors. 
Note that operation $\circ$ is an element-wise product of vectors and $\epsilon=0.25$.
Noise vector $\mathcal{N} \in \R^{N}$ was generated from the multivariate standard normal distribution $\boldsymbol{\mathcal{N}}(\boldsymbol{0},\boldsymbol{I})$ every time the context vector was given.
For every context, these functions have a unique global minimum whose evaluation value is zero.
We set the number of dimensions for the context vector to $N_\alpha=2$ and the number of optimization results for warm starting to $M_\mathrm{prev} = 10$.

Because CMA-ES-CWS uses the past optimization results, we pre-optimized $M_\mathrm{prev}$ problems corresponding to $M_\mathrm{prev}=10$ context vectors generated on $[-2,2]^{N_\alpha}$ uniformly at random.
For these pre-optimizations, we ran CMA-ES with the initial step-size, and covariance matrix was given by $\sigma^{(0)} = 2$ and $\cov^{(0)} = \I$. 
The initial mean vector was given on $[-1,1]^N$ uniformly at random.
The maximum number of evaluations was $1 \times 10^4$ for the sphere and Easom functions, and $4 \times 10^4$ for the Rosenbrock function.
We restarted CMA-ES when the maximum eigenvalue of $\sigma^2 \cov$ was less than $10^{-10}$ and terminated the optimization when the best evaluation value was less than $10^{-8}$ or the number of evaluations reached its maximum value.

The contextual CMA-ES was evaluated using the model obtained in pre-optimization.
In pre-optimization, the context vector was given on $[-2,2]^{N_\alpha}$ uniformly at random before generating each solution.
For a fair comparison with CMA-ES-CWS, the maximum number of evaluations was $M_\mathrm{prev} \times 10^4$ for the sphere and Easom functions and $4 M_\mathrm{prev} \times 10^4$ for the Rosenbrock function.

For each of the objective functions corresponding to $M_\mathrm{prev}$ context vectors, the initialization of WS-CMA-ES was performed with the set of solutions that were generated on $[-2,2]^N$ uniformly at random.
For a fair comparison with CMA-ES-CWS, the number of solutions was $1 \times 10^4$ for the sphere and Easom functions and $4 \times 10^4$ for the Rosenbrock function.
To select a similar task from $M_\mathrm{prev}$ tasks, we measured the Euclid distances between the target and $M_\mathrm{prev}$ context vectors and selected the task corresponding to the nearest context vector to the target context vector.

We also ran CMA-ES for comparison with CMA-ES-CWS.
The initial step-size and covariance matrices were given as $\sigma^{(0)} = 2$ and $\cov^{(0)} = \I$, and the initial mean vector was given on $[-1,1]^N$ uniformly at random.
Target context vector $\context_{\mathrm{new}}$ was given on $[-2,2]^{N_\alpha}$ uniformly at random.
We regarded an optimization as successful when the best evaluation value on the objective function $f(\x; \context_{\mathrm{new}})$ corresponding to this context vector was less than $10^{-8}$.
We performed 20 independent trials for each experimental setting.

\begin{figure*}[!t]
\centering
\includegraphics[width=0.8\hsize]{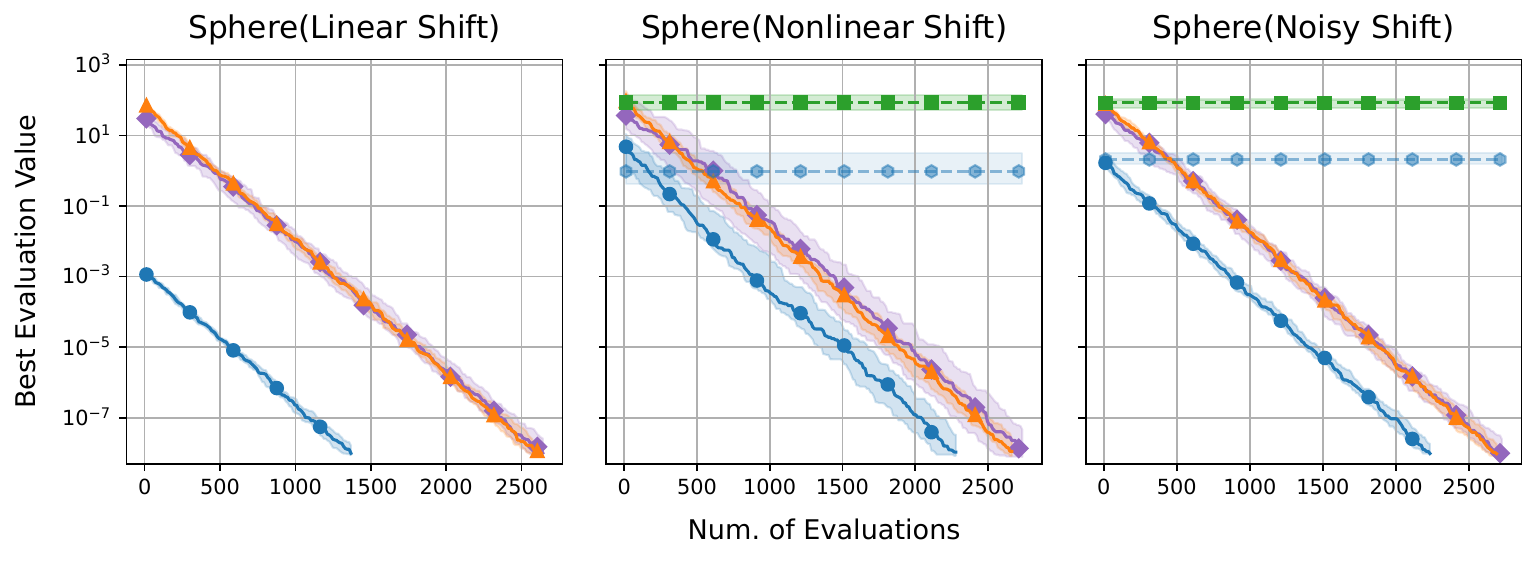} \\
\includegraphics[width=0.8\hsize]{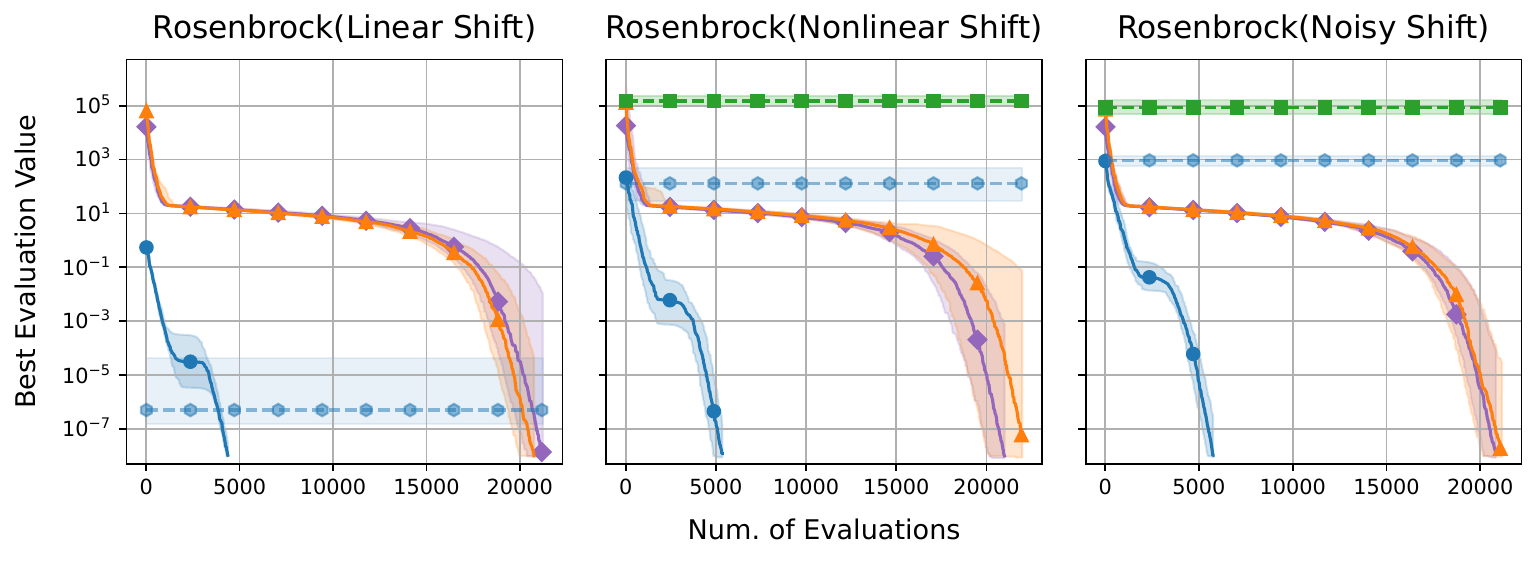} \\
\hspace*{0.16cm}\includegraphics[width=0.94\hsize]{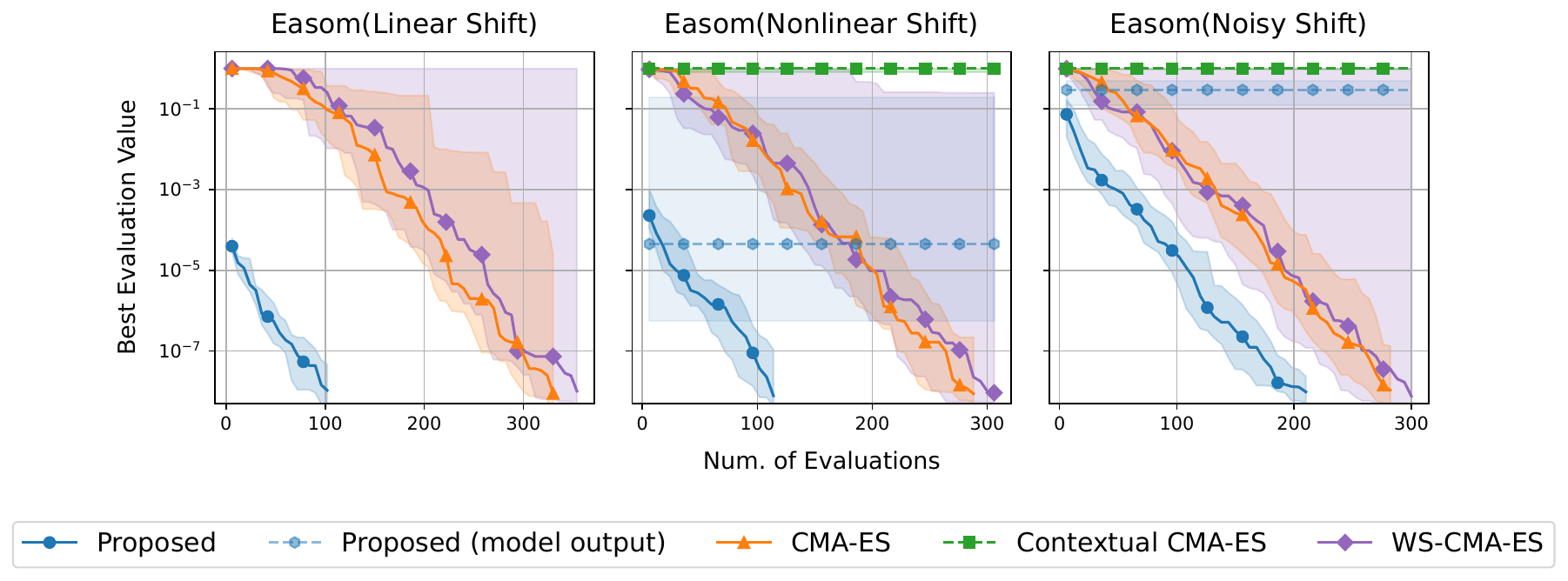}
\caption{
Transitions of best evaluation values on the sphere, Rosenbrock, and Easom functions. 
We plot the medians and interquartile ranges.
Dash lines show the evaluation values for outputs of the policy model in contextual CMA-ES and multi-output GPR in CMA-ES-CWS. Note that the evaluation values less than $10^{-8}$ are not plotted.
}
\label{fig:all10}
\end{figure*}
%
%
%




%
%
%
\begin{figure*}[t]
\centering
\includegraphics[width=0.8\hsize]{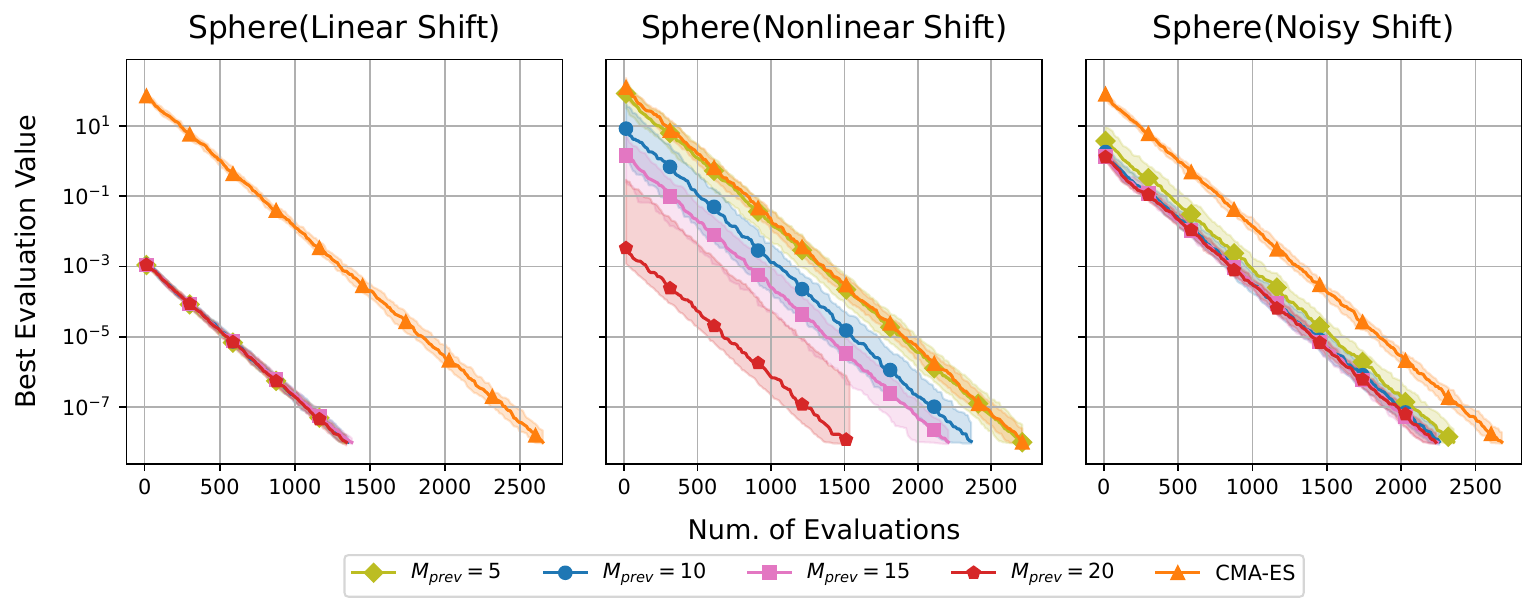}
\caption{Transitions of the best evaluation values with various past optimization results $M_\mathrm{prev}$.
We plotted the medians and interquartile ranges of 20 trials on the sphere function.
}
\label{fig:sphere_abb}
\end{figure*}
%
%
%

\subsection{Experimental Result}\label{sec:experiment:benchmark:comparison}
Figures~\ref{fig:all10} shows the transitions of the best evaluation values on each benchmark function.
To evaluate the multi-output GPR in CMA-ES-CWS, we presented the evaluation value for the mean vector of the predictive distribution.
The contextual CMA-ES and the multi-output GPR in CMA-ES-CWS do not require an additional optimization for the target context vector.
Note that these evaluation values were not plotted for values less than $10^{-8}$.

First, comparing the performance of CMA-ES-CWS with those of CMA-ES and WS-CMA-ES, CMA-ES-CWS consistently outperformed these methods under all setting.
Particularly on the Rosenbrock function, CMA-ES-CWS reached the target evaluation value with approximately $1/4$ of number of the evaluations of CMA-ES and WS-CMA-ES. 
The WS-CMA-ES exhibited no significant advantage over CMA-ES and increased the upper quartile range for the Easom function.
From these results, we confirmed the effectiveness of warm starting in CMA-ES-CWS using the context vector.

Next, comparing the performance of CMA-ES-CWS with that of the contextual CMA-ES, the performance of CMA-ES-CWS in the initial iteration was better than that of the contextual CMA-ES when the nonlinear and noisy shifts were applied.
This is because the contextual CMA-ES utilized a linear model whereas CMA-ES-CWS used the multiple-output GPR that can capture non-linear relationships.
With the linear shift, the initial evaluation value of CMA-ES-CWS was significantly worse than these model outputs.
This is because the lower bound on the initial step-size in~\eqref{eq:proposed:sigma} gave a relatively large initial value when compared with the distance between the initial mean vector and optimal solution.
On the Rosenbrock function with the linear shift, the evaluation value for the mean vector of the predictive distribution was not less than $10^{-8}$ because of the failure of pre-optimizations.

Finally, we compared the results of CMA-ES-CWS for various shift types.
On the sphere function, the linear shift required fewer evaluations than the nonlinear and noisy shifts.
With the Rosenbrock function, the number of evaluations did not change significantly regardless of which transformation was applied.
Note that the CMA-ES-based methods must adapt the covariance matrix to fit the landscape of the Rosenbrock function.
For all transformations, CMA-ES-CWS achieved the initial distribution that required a small number of evaluations for covariance matrix adaptation on the Rosenbrock function.

\subsection{Effect of Number of Pre-optimizations}\label{sec:experiment:benchmark:ablation}
We evaluated the performance of CMA-ES-CWS on the sphere function with various numbers of pre-optimizations $M_\mathrm{prev}$($M_\mathrm{prev} = 5,10,15,20$).
Other experiment settings were the same as in Section~\ref{sec:experiment:benchmark:setting}.
Figure~\ref{fig:sphere_abb} shows the transitions of the best evaluation values.
We also plotted the result of the CMA-ES to consider the case where no pre-optimization result was obtained.
With the linear shift, the optimization performance did not change regardless of the number of pre-optimizations;
because, owing to the lower bound on the initial step-size in~\eqref{eq:proposed:sigma}, the initial step-size was set to a large value such that the performance differences in initial mean vectors were disappeared.
When applying the nonlinear shift, the optimization performance improved as the number of pre-optimizations increased.
Particularly, with $M_\mathrm{prev}=5$, the optimization performance was the same as that of CMA-ES.
For problems with nonlinear dependencies on context vectors, CMA-ES-CWS may be more effective with a larger number of pre-optimization results.
Finally, focusing on the results with the noisy shift, although the optimization performance dropped slightly when $M_\mathrm{prev}=5$, almost the same optimization performance was observed with various numbers of pre-optimizations.
Because the results with the noisy shift were worse than the results with the linear shift, CMA-ES-CWS cannot deal with the noise even with a large number of pre-optimization results.
Developing new noise handling capabilities for warm starting in noisy contextual optimization is a future work.



\begin{figure*}[t]
  \begin{minipage}[b]{0.29\linewidth}
    \centering
    \includegraphics[width=1\hsize]{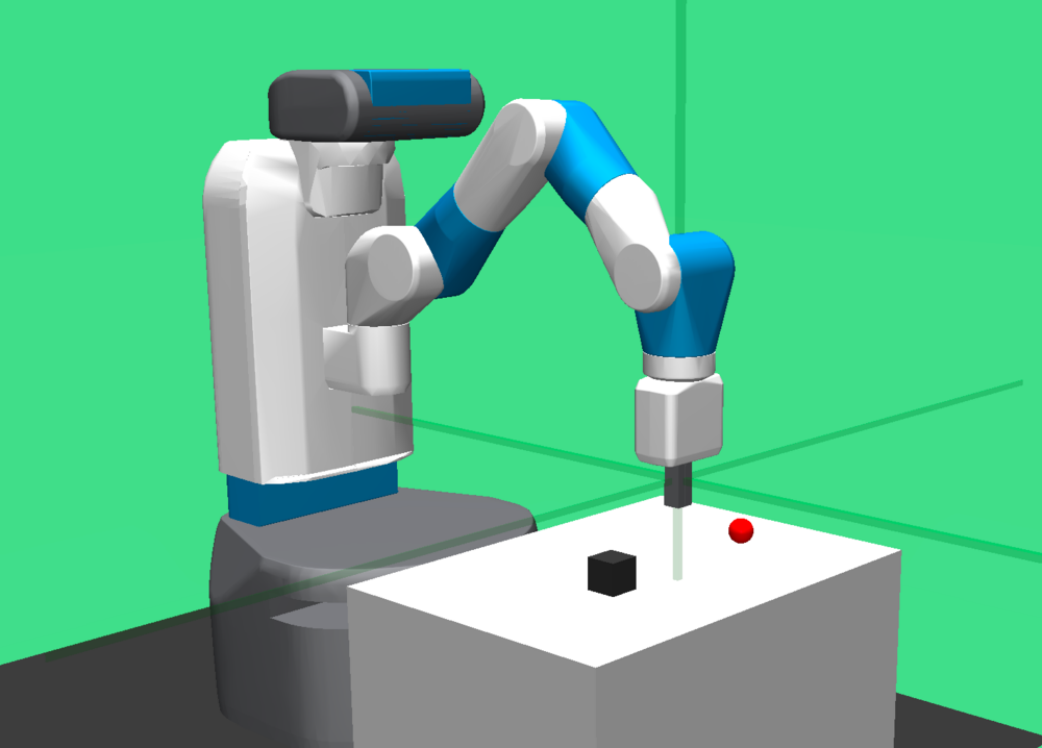}
    \subcaption{FetchPush-v2}
  \end{minipage}
  \begin{minipage}[b]{0.67\linewidth}
    \centering
    \includegraphics[width=0.8\hsize]{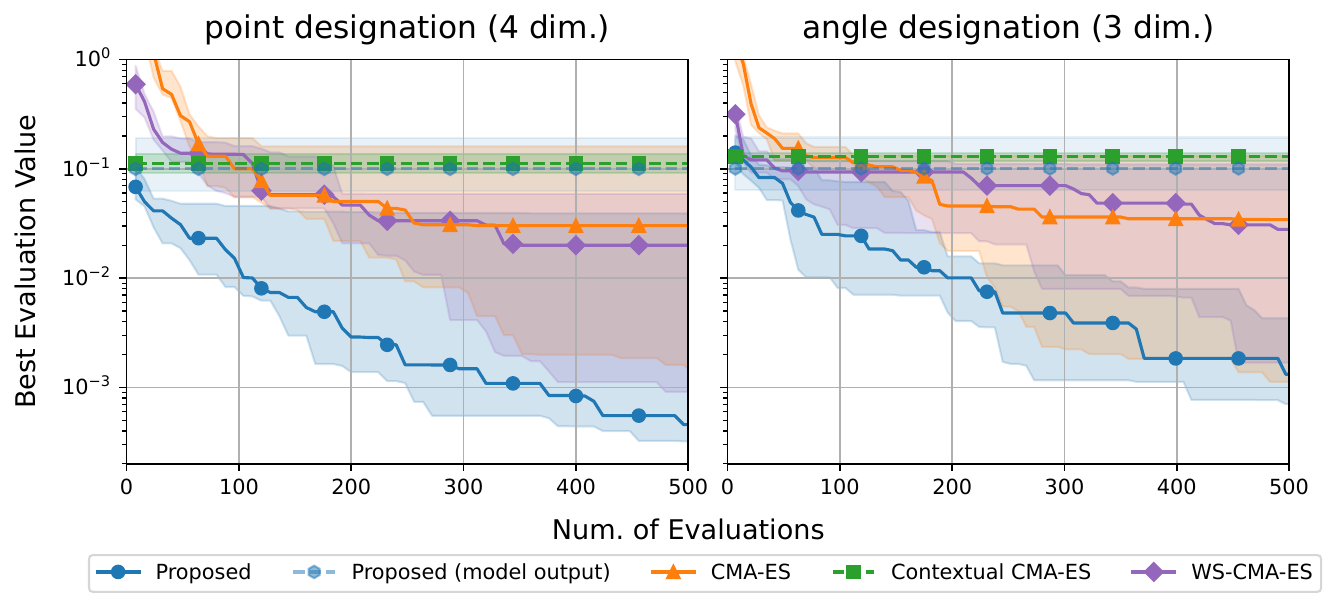}
    \subcaption{Result of algorithm comparison on FetchPush-v2}
  \end{minipage} \\
  \begin{minipage}[b]{0.29\linewidth}
    \centering
    \includegraphics[width=1\hsize]{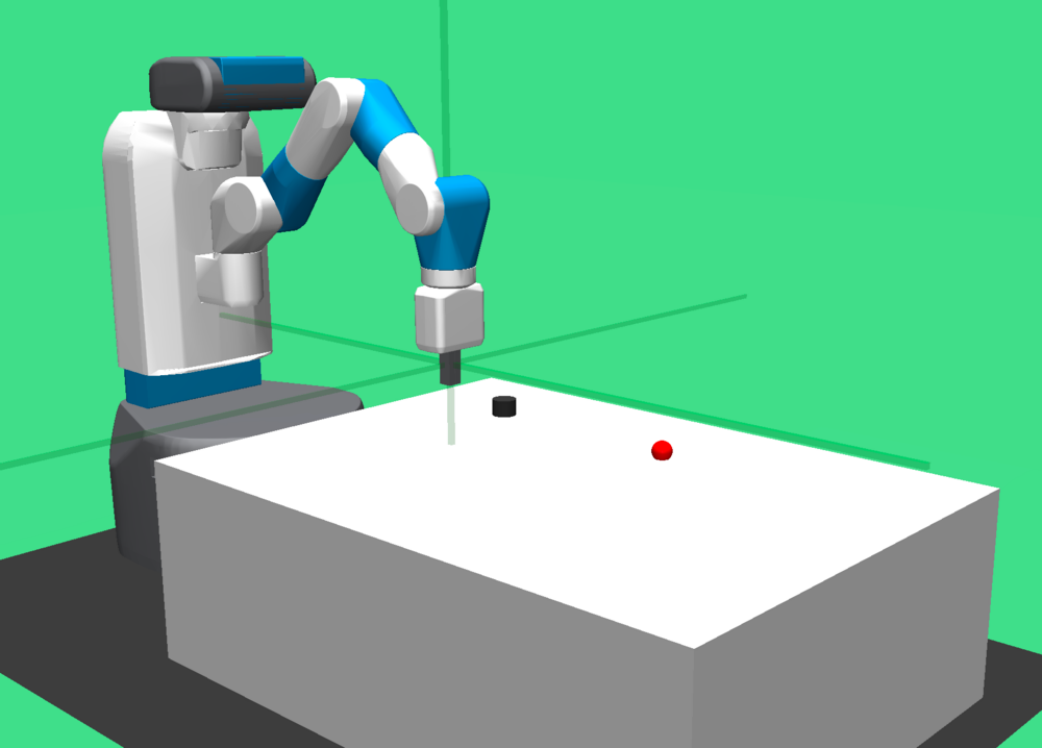}
    \subcaption{FetchSlide-v2}
  \end{minipage}
  \begin{minipage}[b]{0.67\linewidth}
    \centering
    \includegraphics[width=0.8\hsize]{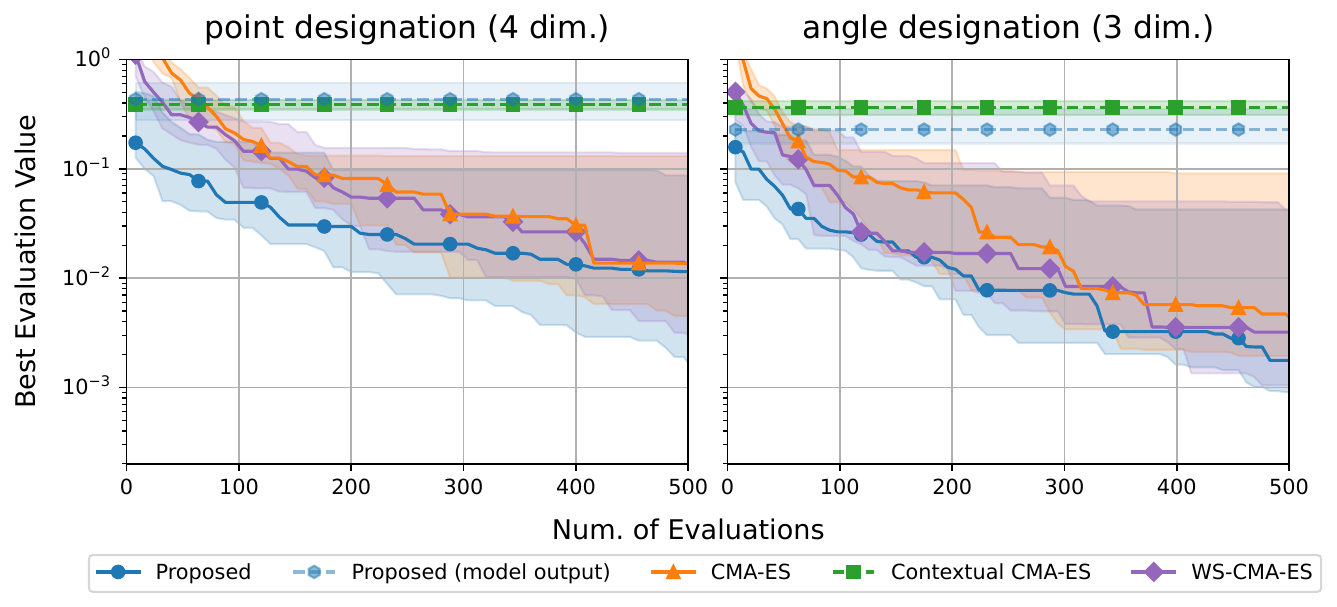}
    \subcaption{Result of algorithm comparison on FetchSlide-v2}
  \end{minipage}
  \caption{Images of FetchSlide-v2 and FetchPush-v2 and comparison results}
  \label{fig:fetch:result}
\end{figure*}

\section{Evaluation Experiment in Robot Control Task}

\subsection{Experimental Setting}

We used two robot control tasks provided by OpenAI Gym~\cite{openai,multigoal}, 
\begin{itemize}
    \item \textbf{FetchPush-v2}: Control the robot arm to push a box to a target position.
    \item \textbf{FetchSlide-v2}: Control the robot arm to slide a box to a target position that is out of reach for the arm.
\end{itemize}
Following reference~\cite{Pinsler:2019}, we considered a trajectory parameter space to reduce the number of dimensions of the problem.
The objective of optimization was to designate the parameters that determine the trajectory of the arm.
The trajectory comprised two movements: putting the arm close to the box and sliding the arm to move the box.
The trajectory was determined using coordinates $\x_1, \x_2 \in \R^{2}$.
The first point $\x_1$ determined where to put the arm, whereas the second point $\x_2$ determined where to slide the arm from the first position to push or slide the box.
The maximum amount of movement for each time step was set to (1,1,1), whereas OpenAI Gym set time step to 1/25th of a second.
The simulator calculated the arm trajectory with the above limits.

We prepared two trajectory parameter spaces: point and angle designation spaces.
In point designation space, there were four design variables that determined points $\x_1$ and $\x_2$.
The ranges of design variables were set as $\x_1,\x_2 \in [-0.2,0.2]^2$ for FetchPush and $\x_1\in [-0.2,0]\times[-0.2,0.2]$ and $\x_2 \in [0,0.4]\times[-0.4,0.4]$ for FetchSlide.
The angle designation space contained three design variables, point $\x_2$, and additional parameter $\theta$ that determined the angle for approaching to the box.
The point $\x_1$ was set on the circle with the radius 0.07 around the initial box position and determined using parameter $\theta$.
The ranges of the three design variables were set as $\theta \in [0,2\pi]$ and $ \x_2 \in [-0.2,0.2]^2$ for FetchPush and $\theta \in [0,\pi]$ and $\x_2 \in [0,0.4]\times[-0.4,0.4]$ for FetchSlide.
We used the final distance between the box and the target position as the evaluation value in each task.
When the design variable was out of the range, we computed the evaluation value using the design variable clipped into the range and, as a penalty, added the sum of the Euclidean distances from the range to the evaluation value.

In this experiment, we used four-dimensional context vector $\context \in \R^{4}$.
The first two dimensions determined the initial position of the box and other dimensions determined the target position.
We set the range of the context vector to $[-0.15,0.15]^4$ for FetchPush and $[-0.1,0.1]^2\times [-0.3,0.3]^2$ for FetchSlide.

The number of pre-optimizations was $M_\mathrm{prev} = 10$.
The maximum number of evaluations in each pre-optimization was $5\times10^2$ for CMA-ES-CWS, CMA-ES, and WS-CMA-ES.
For a fair comparison, we set the maximum number of evaluations for contextual CMA-ES to $5M_\mathrm{prev}\times10^2$.
Other experimental settings were the same as in the experiment using benchmark functions in Section~\ref{sec:experiment:benchmark}.



\subsection{Experimental Result}
Figure~\ref{fig:fetch:result} shows the transitions of the best evaluation values for the two tasks.
Note that these best evaluation values included penalty, and we confirmed that only CMA-ES and WS-CMA-ES included penalty until approximately 100 evaluations, whereas other methods did not include the penalty.

In all settings, CMA-ES-CWS consistently outperformed other methods.
Particularly, in the FetchPush task, CMA-ES-CWS significantly performed better than did the other methods.
It is evident that CMA-ES and WS-CMA-ES started optimizations from out of the range for the design variables, and these methods consumed evaluations to generate solutions in the range.
By contrast, CMA-ES-CWS could start the optimization from in the range for the design variables.
We can confirm that the warm starting of CMA-ES-CWS was also effective in robotic tasks.

\section{Conclusion}
In this study, we proposed CMA-ES-CWS, which incorporates a warm starting for contextual optimization problems.
The CMA-ES-CWS uses a multi-output GPR to compute the predictive distribution of the optimal solution corresponding to a newly given context vector from the best solutions corresponding to previously given context vectors. 
Subsequently, it initializes the sampling distribution of CMA-ES using the predictive distribution. 
Based on experimental results using benchmark functions, we confirmed that CMA-ES-CWS exhibited better performance than that of the contextual CMA-ES and WS-CMA-ES when applying nonlinear or noisy transformations of the search space.
In addition, the CMA-ES-CWS outperformed also the other methods in the robot control tasks.
In future, we will extend CMA-ES-CWS to discrete and mixed-integer optimization methods by introducing Bayesian estimation using discrete probability distributions.
The development of a reasonable initialization method for the covariance matrix is also planned.


\bibliographystyle{splncs04}
\bibliography{reference}

\end{document}